\documentclass{article}

\usepackage[preprint]{neurips_2026}

\usepackage[utf8]{inputenc} 
\usepackage[T1]{fontenc}    
\usepackage{hyperref}       
\usepackage{url}            
\usepackage{booktabs}       
\usepackage{amsfonts}       
\usepackage{nicefrac}       
\usepackage{microtype}      
\usepackage{graphicx}
\usepackage{amsmath}
\usepackage{cleveref}
\usepackage{subcaption}
\usepackage{wrapfig}
\usepackage{todonotes}
\usepackage{colortbl}
\usepackage[table]{xcolor}
\usepackage{multirow}
\usepackage{amssymb}
\usepackage{tikz}
\newcommand{\Ldownarrow}{%
  \begin{tikzpicture}[baseline=-0.2em]
    \draw[->,>=stealth] (0,0.5em) -- (0,0em) -- (1.0em,0);
  \end{tikzpicture}%
}
\definecolor{elegantblue}{HTML}{0077FF}
\hypersetup{
    colorlinks=true,
    linkcolor=elegantblue,
    urlcolor=elegantblue,
    citecolor=elegantblue
}

\title{Orbis 2: A Hierarchical World Model for Driving}

\author{%
  Sudhanshu Mittal\thanks{Main Contributors. \texttt{\{mittal,mousakha,galessos\}@cs.uni-freiburg.de}\\} \\
  \And
  Arian Mousakhan\footnotemark[1] \\
  \And
  Silvio Galesso\footnotemark[1] \\
  \And
  Karim Farid \\
  \AND
  Johannes Dienert \\
  \And
  Rajat Sahay \\
  \And
  Muhammad Ali \\
  \And
  Thomas Brox \\
}

\begin{document}

\maketitle

\begin{center}\vspace{-2.00em}
     University of Freiburg, Germany
\end{center}

\vspace{2em}
\begin{abstract}

Current world models operate at a single level of abstraction, with most prioritizing perceptual fidelity while lacking the spatial reasoning and semantic understanding required for real-world downstream tasks.
We present a hierarchical driving world model that factorizes future prediction across two levels operating at distinct temporal and abstraction scales: a high-level predictor that forecasts coarse scene structure over extended temporal horizons, and a low-level generator that produces detailed predictions conditioned on the high-level output.
This decomposition yields high perceptual fidelity while also capturing strong spatial and semantic representations.
We further show that pretraining with a diffusion forcing objective yields substantially richer internal representations than the standard teacher forcing objective, while teacher forcing — predicting only the next frame from clean context — produces more stable autoregressive rollouts.
We therefore introduce a generic two-stage training paradigm that pretrains the model with diffusion forcing and fine-tunes with teacher forcing, combining the representational benefits of the former with the rollout stability of the latter.
Our approach achieves state-of-the-art results across the standard suite of driving world model evaluations on established benchmarks, including long-horizon generation fidelity, steering responsiveness evaluated on counterfactual scenarios, and internal representation quality. Project page with code, demo, checkpoints and qualitative results: \href{https://lmb-freiburg.github.io/orbis2.github.io}{https://lmb-freiburg.github.io/orbis2.github.io}

\end{abstract}

\section{Introduction}

\begin{figure}[t]
    \centering
    \includegraphics[width=\linewidth]{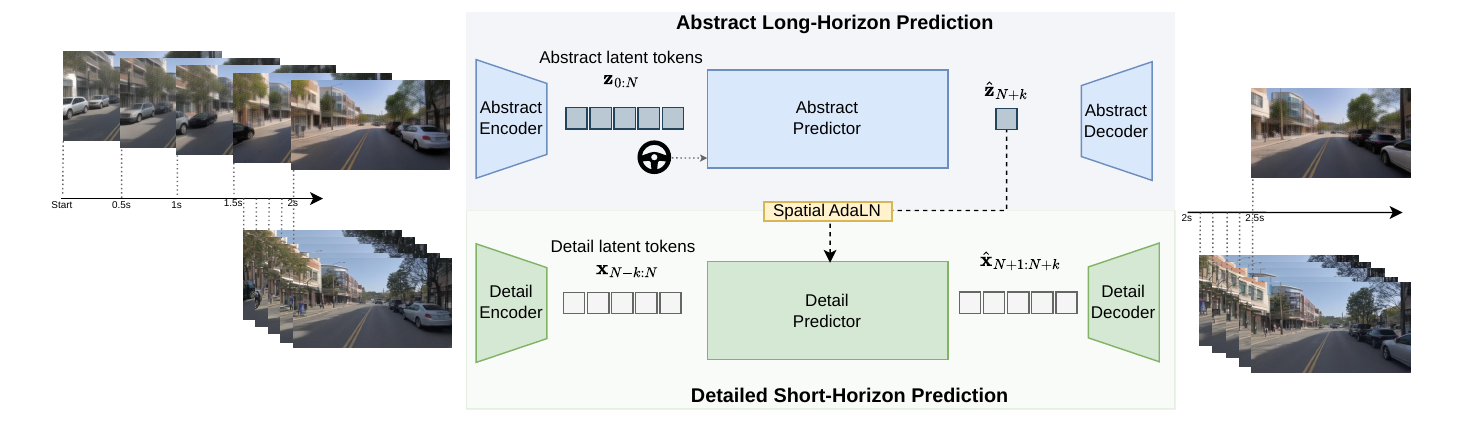}
    \caption{
        \textbf{Orbis 2: a hierarchical driving world model.}
        The \emph{abstract predictor} operates over a long temporal context to forecast a future state in latent space, capturing abstract scene dynamics over long horizons, and enabling steering control.
        The \emph{detail predictor} is conditioned on this abstract prediction and generates fine-grained short-horizon frames, enabling high-fidelity local prediction grounded in long-range temporal context.
    }
    \label{fig:framework}
\end{figure}

World models~\cite{ha2018world, hafner2023mastering, LeCun2022APT} are learned dynamical systems that approximate how an environment evolves in response to an agent's actions, providing internal predictions of transition dynamics that support planning, simulation, and policy learning~\cite{sutton1981adaptive}.
In high-dimensional, complex domains such as autonomous driving, world models are commonly formulated in a learned latent space where an encoder maps raw observations to compact latent representations and a predictor propagates these states forward, conditioned on the agent's actions.
Current implementations of this paradigm are limited in two key respects.
First, the predictor operates at a single temporal scale~\cite{bar2025navigation, assran2025v}: rollouts are generated frame-by-frame, accumulating error over time and limiting the model's ability to explicitly reason about long-horizon scene evolution.
Second, the latent space is typically shaped by a reconstruction objective that preserves all fine visual details such as texture and appearance, even though long-horizon dynamics are primarily governed by coarse, slowly varying semantic properties, making such representations suboptimal for long-horizon forecasting.
This aligns with a well-established principle in cognitive science: humans plan over distant goals using simplified, abstract representations and refine details only as actions approach execution~\cite{Tomov499418, ho2022people}.
Building on this principle, we propose a setup that decomposes future prediction across two levels of abstraction.
A high-level predictor forecasts long-horizon dynamics in an abstract latent space, producing subgoals that condition a low-level predictor. The latter operates on a finer-grained latent space, specializing in detailed, short-horizon prediction. \Cref{fig:framework} provides an overview of our hierarchical framework.

For the high-level predictor, we require a latent space that strongly captures spatial structure and semantic understanding, enabling long-horizon temporal estimation. We therefore utilize a compressed DINO~\cite{oquab2023dinov2} representation as the target for our high-level predictor. 
Since directly applying flow matching~\cite{lipman2022flow} in the high-dimensional space is challenging~\cite{zheng2025diffusion, kumar2026learning}, we introduce a lightweight projection that compresses DINOv2 features into a lower-dimensional latent space, which stabilizes training and improves long-horizon rollout quality, while retaining the representation properties of DINO. The goal of the low-level predictor is then to capture fine-grained details and precise motion dynamics, conditioned on the high-level prediction. This is naturally suited to pixel-reconstruction based latent spaces such as VAEs~\cite{kingma2013auto}, which preserve the fine-grained visual detail needed for short-horizon, high-fidelity generation.

To implement the two predictors, we rely on flow matching~\cite{lipman2022flow}, which was shown to be particularly effective for long-horizon prediction in driving world models~\cite{mousakhan2026overcoming}.
As the major objective of world models is to generalize to unseen scenarios observed previously only in different contexts, an important indicator of the quality of a world model is the quality of its learned internal representations. Driving world models have typically been evaluated with FVD~\cite{NEURIPS2024_a6a066fb,agarwal2025cosmos, mousakhan2026overcoming}, which measures the visual fidelity of generated frames. The high-level predictor operating in an abstract latent space, however, is not expected to reproduce fine-grained appearance, even though it may capture the scene structure that matters for long-horizon prediction.
Thus, we additionally evaluate the quality of the internal representations of both predictors via linear probing on semantic segmentation and depth estimation, which directly measures how well the learned latent space encodes scene structure and geometry. In the proposed hierarchical world model, both predictors can play to their advantages: the abstract predictor learns higher-quality structural representations and dynamics, whereas the low-level predictor can satisfy the requirements of high-fidelity, short-horizon roll-outs with strong FVD scores.
By probing the internal representations of the model, we find that the standard teacher forcing recipe — conditioning on clean ground-truth context to predict the target frame — fails to learn strong internal representations, as the model never observes out-of-distribution context during training.
Pretraining with a \emph{diffusion forcing} objective~\cite{chen2024diffusion}, which independently corrupts and denoises all frames in the sequence, produces substantially richer internal representations. We summarize our contributions as follows:

    \textbf{A hierarchical driving world model.} We introduce a two-level world model that decomposes future prediction into a high-level predictor operating in an abstract latent space and a low-level predictor producing high-fidelity frames.
 
    \textbf{A two-stage training paradigm for predictive representation learning.} We propose pretraining with a diffusion forcing objective followed by short fine-tuning with teacher forcing. This combination enables self-supervised representation learning directly from the temporal prediction loss and substantially improves representation quality over only next frame prediction objective.

    \begin{wrapfigure}{r}{0.47\textwidth}
      \vspace{-0.5em}
      \centering
      \includegraphics[width=\linewidth]{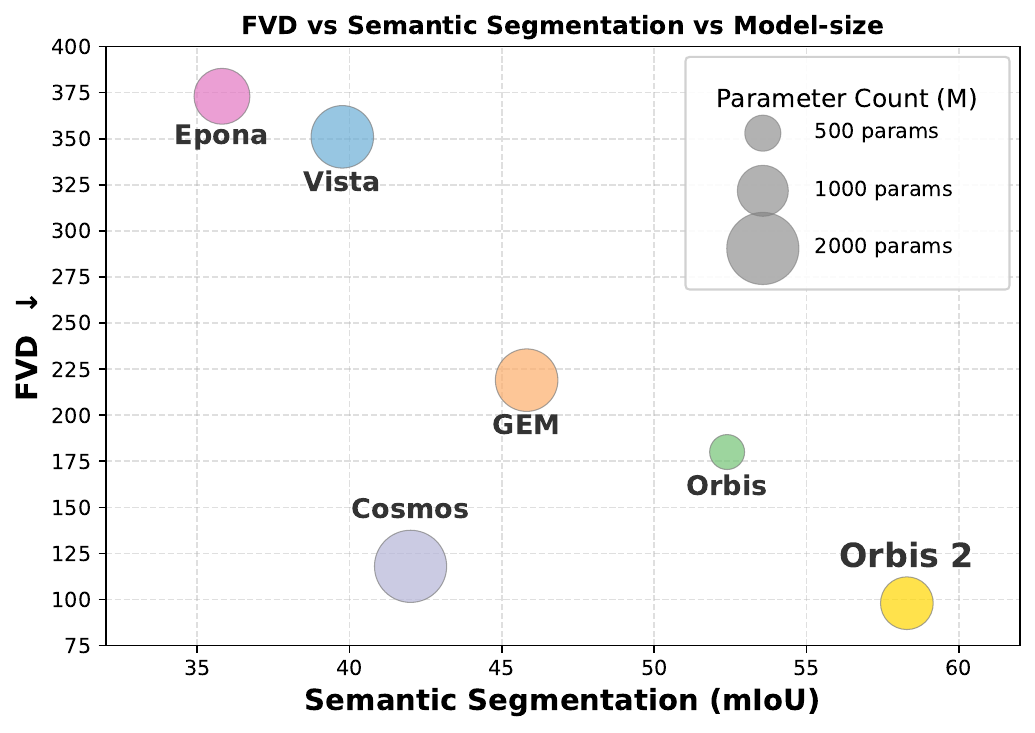}
        \caption{Comparison of driving world models over model scale (parameters), FVD (6s rollouts) and representation quality based on probing for semantic segmentation. FVD evaluated on Waymo and segmentation on Cityscapes.
        \vspace{-3em}
        }
      \label{fig:bubblechart}
    \end{wrapfigure}

    \textbf{State-of-the-art results on driving world modeling.} As shown in Fig.~\ref{fig:bubblechart}, our model achieves state-of-the-art performance on video-based driving world modeling across complementary axes: long-horizon perceptual fidelity (FVD) and representation quality~(linear probing for semantic segmentation and depth). Our model also outperforms prior methods in steering responsiveness on counterfactual scenarios.

\section{Related Work}

World models for autonomous driving aim to predict future scene evolution conditioned on past observations and ego actions.
A first line of work models this prediction in \emph{discrete token space}: GAIA-1~\cite{hu2023gaia}, DrivingWorld~\cite{hu2024drivingworld}, and DrivingGPT~\cite{Chen_2025_ICCV} employ autoregressive transformers over quantized visual tokens, while Cosmos~\cite{agarwal2025cosmos} scales this paradigm to large-scale video pretraining.
A second line operates in \emph{continuous latent space}, typically with diffusion~\cite{ho2020denoising, karras2022elucidating} or flow-matching~\cite{lipman2022flow} objectives: Vista~\cite{NEURIPS2024_a6a066fb}, DriveDreamer~\cite{wang2024drivedreamer}, GenAD~\cite{yang2024genad}, MagicDrive-V2~\cite{Gao_2025_ICCV}, Epona~\cite{zhang2025epona}, and Orbis~\cite{mousakhan2026overcoming} demonstrate high-fidelity multi-frame generation across diverse driving scenarios.
Orbis directly compares these two paradigms under a hybrid tokenizer and finds that continuous representations are less brittle and more effective for long-horizon prediction than discrete tokens.
Despite differing in their representation choices, all of these models share a single level of abstraction in their predictor.

\subsection{Hierarchical planners}
In reinforcement learning, hierarchical approaches have been a popular paradigm to learn both high-level skills and detailed control~\cite{SUTTON1999181,Li2020Sub-policy}. Director~\cite{hafner2022deep} learns a two-level hierarchy where a high-level planner selects subgoals in latent space and a low-level policy executes them.
Many recent vision-language-action models can also be regarded as hierarchical planners, with language-based planning covering the abstract, strategic level, whereas a fine-grained policy covers low-level control. For example, Hi~Robot~\cite{shi2025hi} interprets complex instructions with a VLM to determine the next step, which conditions a low-level action policy.
HiLAM~\cite{kim2026hierarchical} discovers latent skills in an unsupervised fashion by modeling long-term temporal structure, then uses them to condition short-horizon prediction.

\subsection{World models with abstract latent spaces}
THICK~\cite{gumbsch2024learning} learns multi-scale temporal abstractions in the context of hierarchical model-based RL. This is closest to our work, as it includes multiple abstractions on the side of the predictor.
DINO-WM~\cite{zhou2024dino} does not implement a hierarchy, but uses DINO features as latents for its predictor, which is similar to our abstract-level predictor.
In concurrent work, Zhang et al.~\cite{zhang2026hierarchical} use predictors at two levels for hierarchical planning, but both predictors share the same latent space. They demonstrate that planning with this hierarchy of predictors substantially improves the success rate compared to various flat models.

\section{Method}

\subsection{Preliminary: Flow Matching}
We adopt the flow matching objective~\cite{lipman2022flow} to model the distribution of future frames $\mathbf{x}_N$ in latent space.
A noisy interpolant is constructed as $\mathbf{x}_N^\tau = (1-\tau)\mathbf{x}_N + \tau\epsilon$, where $\epsilon \sim \mathcal{N}(0, I)$ and $\tau \in [0,1]$, defining a linear path from data to noise.
A velocity network $\mathbf{v}(\mathbf{x}_N^{\tau};\, \mathbf{x}_{0:N-1})$, conditioned on context frames $\mathbf{x}_{0:N-1}$, is trained to regress the ground-truth velocity field:
\begin{equation}
    \mathcal{L}_{\mathrm{FM}} = \mathbb{E}_{\tau,\,\epsilon}
    \big[\|\mathbf{v}(\mathbf{x}_N^{\tau};\, \mathbf{x}_{0:N-1}) - (\epsilon - \mathbf{x}_N)\|^2 \big].
\label{eq:flow_loss}
\end{equation}
During inference, the predicted frame is generated by integrating the learned velocity field from $\tau{=}1$ to $\tau{=}0$, initialized from $\mathbf{x}_N^1 \sim \mathcal{N}(0,I)$.

\paragraph{Teacher forcing and diffusion forcing:}
A central design choice is how the velocity network conditions on context frames during training.
Under \textbf{teacher forcing}, context frames are always provided as 
clean ground-truth observations and the velocity is predicted only for 
the last frame, as in \cref{eq:flow_loss}.
In contrast, \textbf{diffusion forcing}~\cite{chen2024diffusion} 
independently corrupts each frame $\mathbf{x}_i$ in the full sequence 
$\mathbf{x}_{0:N}$ with an independent noise level $\tau_i \sim [0,1]$, and 
trains the model to predict the velocity for \emph{all} frames jointly:
\begin{equation}
    \mathcal{L}_{\mathrm{DF}} = \mathbb{E}_{\tau_{0:N},\,\epsilon_{0:N}}
    \big[\|\mathbf{v}(\mathbf{x}_{0:N}^{\tau_{0:N}}) - (\epsilon_{0:N} - \mathbf{x}_{0:N})\|^2 \big].
\label{eq:df_loss}
\end{equation}
While teacher forcing has emerged as a common training objective for video world models~\cite{mousakhan2026overcoming, agarwal2025cosmos, zhang2025epona, valevski2024diffusion},
we find that it yields weaker internal representations, as the model is never exposed to corrupted context during training.
We demonstrate that diffusion forcing leads to stronger internal representations.

\subsection{Hierarchical World Model}
\label{sec:hierarchical_wm}

We propose a two-level hierarchical world model that decomposes future prediction along both abstraction and temporal axes.
The high-level predictor $\mathcal{F}_H$ operates at a coarse temporal resolution in an abstract latent space, capturing long-horizon dynamics. The low-level predictor $\mathcal{F}_L$ operates at a finer temporal resolution in a pixel-aligned latent space, making detailed short-horizon prediction conditioned on the high-level prediction.
An overview of the architecture is shown in \Cref{fig:framework}.

\paragraph{Tokenizer latent representations:} Let $o$ denote an observed driving frame. 
For each frame $o$, we extract two complementary representations.
The fine-grained latent $\mathbf{x} = \mathcal{E}_L(o)$ is produced by a tokenizer encoder $\mathcal{E}_L$ trained with a reconstruction objective~\cite{rombach2022high, esser2021taming}. This latent preserves pixel-aligned details and is used by the low-level predictor.
In parallel, we learn an abstract encoder $\mathcal{E}_H$ that produces a high-level latent representation, 
$\mathbf{z} = \mathcal{E}_H(o).$ Unlike the detail tokenizer, $\mathbf{z}$ is optimized to capture semantic and structural information useful for long-horizon prediction. We train $\mathcal{E}_H$ by aligning its representation with a frozen DINO embedding~\cite{oquab2023dinov2} through a learned projection head
$\mathcal{P}$. We add a weak reconstruction regularizer through a lightweight decoder $\mathcal{D}_H$,
weighted by $\lambda_{\mathrm{rec}}=0.1$, so that the abstract latent retains task-relevant visual information without degrading its semantic alignment.
Formally, given an observation $o$ and its frozen DINO embedding $\mathcal{E}_{\mathrm{DINO}}(o)$, we train $\mathcal{E}_H$, $\mathcal{P}$, and $\mathcal{D}_H$ with:
\begin{equation}
    \mathcal{L}_{\mathrm{abs}} \;=\; \underbrace{\big\| \mathcal{P}\big(\mathcal{E}_H(o)\big) - \mathcal{E}_{\mathrm{DINO}}(o) \big\|^2}_{\text{DINO alignment}} \;+\; \lambda_{\mathrm{rec}} \, \underbrace{\big\| \mathcal{D}_H\big(\mathcal{E}_H(o)\big) - o \big\|^2}_{\text{reconstruction regularizer}},
\label{eq:abstract_encoder_loss}
\end{equation}
The resulting abstract latent $\mathbf{z}$ is only $16$-dimensional latent space,
providing a compact DINO-aligned representation that is well suited to the flow-matching
objective of the high-level predictor.

\paragraph{Abstract long-horizon predictor:}
The high-level predictor $\mathcal{F}_H$ models future dynamics in the abstract latent space at a coarse temporal resolution.
Let $\mathbf{z}_i$ denote the abstract latent of the frame at time step $i$, and let $\mathbf{Z}_N^k = \{\mathbf{z}_0, \mathbf{z}_k, \mathbf{z}_{2k}, \ldots, \mathbf{z}_N\}$ denote the strided context of past abstract latents sampled every $k$ frames.
Conditioned on this context and the current ego action $a_N$, $\mathcal{F}_H$ predicts the abstract latent $k$ steps into the future:
\begin{equation}
    \hat{\mathbf{z}}_{N+k} \;=\; \mathcal{F}_H\big(\mathbf{Z}_N^k,\, a_N\big).
\end{equation}

\paragraph{Detailed short-horizon predictor:}
The low-level predictor $\mathcal{F}_L$ operates at fine temporal resolution and is responsible for generating the $k$ intermediate fine-grained frames between $\mathbf{z}_N$ and $\hat{\mathbf{z}}_{N+k}$.  
Let $\mathbf{x}_i$ denote the fine-grained latent at time step $i$.
Conditioned on its own short high-frequency context window of $m$ frames and the abstract subgoal $\hat{\mathbf{z}}_{N+k}$, $\mathcal{F}_L$ predicts the next $k$ frames:
\begin{equation}
    \hat{\mathbf{x}}_{N+1:N+k}
    =
    \mathcal{F}_L\big(
        \mathbf{x}_{N-m:N},\,
        \hat{\mathbf{z}}_{N+k}
    \big).
\end{equation}

During training, the abstract high-level predictor is pre-trained using the diffusion forcing loss~\cref{eq:df_loss} and fine-tuned by teacher forcing loss~\cref{eq:flow_loss}.
The detailed low-level predictor is trained directly with a teacher-forcing objective conditioned on the high-level predictor output.

\paragraph{Action-conditioning:}
We first pre-train the hierarchical predictors unconditionally, without action information.
We then fine-tune the world model to enable conditional generation on planar trajectories. The embedded steering vector is added to the $\tau$ embedding, and fed to the predictor via adaptive layer normalization layers~\cite{perez2018film,peebles2023scalable,bar2025navigation}. We only condition the top-level predictor branch.

\subsection{Diffusion-forcing pre-training for improved predictor representations}
Diffusion forcing~\cite{chen2024diffusion} was originally introduced to improve rollout length by mitigating exposure bias in sequential generation: models trained with teacher forcing observe clean ground-truth context during training, but must condition on their own prediction at inference time. This typically causes error to compound over long rollouts. We discover that diffusion forcing also substantially improves the internal representations learned by the predictor.
Building on this finding, we propose using diffusion forcing as a pre-training stage followed by teacher-forcing fine-tuning for the predictor training. The fine-tuning stage aligns the predictor with the downstream future-prediction objective while retaining the representation quality gained during diffusion forcing. 

We validate this effect through linear probes on semantic segmentation and depth estimation, and observe consistent gains across predictors operating in both detail and abstract tokenizer latent spaces, as well as across both probing tasks. This training strategy is independent of our hierarchical architecture and generalizes to any predictor level. This new training strategy also results in much
stable long-horizon rollouts.  In our model, we apply it to the abstract predictor $\mathcal{F}_H$, where it yields stronger high-level dynamics representations and better downstream world modeling performance.

\section{Experiments}

\subsection{Datasets and Training Details}~\label{sec:datasets}
\Cref{tab:dataset} provides an overview of the training data used in training both predictors. Both levels are trained for a single epoch over the full 5{,}890 hours of video.
Both tokenizers are trained on a curated mix of 2.6 million frames sampled from BDD100K~\cite{yu2020bdd100k}, OpenDV~\cite{yang2024genad}, Honda HAD~\cite{kim2019CVPR}, Honda HDD~\cite{ramanishka2018CVPR}, ONCE~\cite{mao2021one}, nuScenes~\cite{Caesar_2020_CVPR}, and nuPlan~\cite{caesar2021nuplan}.
For steering-conditional generation, the model is fine-tuned on a 500h sub-set of the NVIDIA PhysicalAI AV dataset~\cite{physicalai_av} using the dataset's IMU annotations to compute planar trajectory inputs.
The video dataset~\cite{natix2026_multi_camera_driving_dataset} used in this study was acquired in partnership with NATIX. The dataset is partially publicly available. It is a large-scale, crowd-sourced collection of multi-camera driving data spanning the United States, Europe, and Japan, covering varied road conditions, traffic environments, and regional driving behaviors. For this work, we used a 2,374-hour subset of front-camera driving video data. In future work, we plan to extend this approach using multi-camera data.

The high-level predictor operates at 2Hz with a window of 6 frames (3 seconds), taking 5 frames as input and predicting 1 frame. Its predicted abstract latent serves as the subgoal that conditions the low-level predictor. The low-level predictor operates at 10 Hz with a window of 10 frames (1second), taking 5 frames as context and generating the next 5 frames. During training, the low-level predictor is conditioned on ground-truth abstract latents. At inference, it uses the predicted latents from the high-level predictor. Rollouts are performed autoregressively: the high-level predictor advances with a sliding window of 1 frame and the low-level predictor with a sliding window of 5 frames.
Both predictor models are trained at a resolution of $512 \times 288$. 
\begin{table}[h]
  \small
  \centering
  \caption{Overview of the training datasets used for the world model.*partially public}
  \label{tab:dataset}
  \begin{tabular}{lccccc}
      \toprule
      & OpenDV~\cite{yang2024genad} & BDD100k~\cite{yu2020bdd100k} & PhysicalAI-AV~\cite{physicalai_av} & NATIX~\cite{natix2026_multi_camera_driving_dataset}*  & Total \\
      \midrule
      Hours (h)  & 1124 & 659 & 1633 & 2374 & 5890 \\
      \bottomrule
  \end{tabular}
\end{table}


\subsection{Model design}~\label{sec:model_design}

For the first stage, we use the hybrid tokenizer proposed in~\cite{mousakhan2026overcoming}, consisting of a ViT-based~\cite{dosovitskiy2020image} image encoder with a vector-quantized bottleneck for regularization, trained with a VQGAN~\cite{esser2021taming} objective. The encoder contains 86M parameters, while the decoder is a CNN-based architecture with 91M parameters.
In the hierarchical setup we train a separate tokenizer per predictor. The high-level one targets DINOv2B embedding reconstruction with a small pixel reconstruction weight \mbox{$\lambda_{\mathrm{rec}} = 0.1$}; the low-level one follows the pixel-aligned VQGAN~\mbox{\cite{esser2021taming}} objective.

Following~\cite{mousakhan2026overcoming}, we extend the DiT block~\cite{peebles2023scalable} into a Spatio-Temporal Transformer (ST-Transformer) backbone with 512M parameters. The low-level detail predictor adds 43M parameters through spatially aligned conditioning via adaLN, summing to 555M parameters. The full hierarchical predictor totals 1067M parameters. We maintain an exponential moving average (EMA) of the model weights with a decay of $0.9999$ and use the EMA model at inference. 
For the model-internal design studies, we train at a lower resolution of $336 \times 192$ with a smaller 257M abstract predictor and 279M detail predictor.

\subsection{Evaluation}
\begin{wrapfigure}{r}{0.45\textwidth}
  \vspace{-\intextsep}
  \centering
  \includegraphics[width=\linewidth]{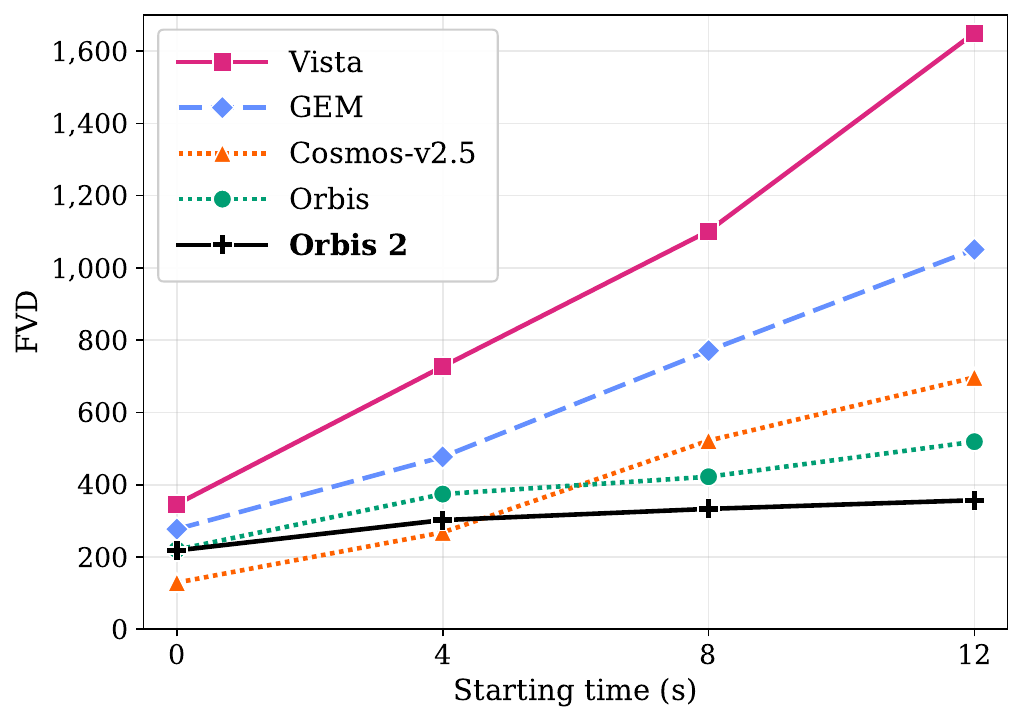}
    \caption{
    Performance (FVD) of long-horizon rollouts on nuPlan-turns over consecutive 4s windows.
    Models with good initial video quality can quickly degrade over time.
    FVD-slope captures this progressive fidelity loss.}
  \label{fig:fvd_chunked}
  \vspace{-1em}
\end{wrapfigure}

\textbf{Representation quality.}
We evaluate internal representations via linear probing on two downstream tasks: semantic segmentation on Cityscapes~\cite{cordts2016cityscapes}, reporting mIoU, and depth estimation on KITTI~\cite{Geiger2012CVPR}, reporting RMSE. Details about linear probing setup are included in Appendix~\ref{app:linear_probe_details}.

\textbf{Predictive performance.}
We assess the visual quality and diversity of predicted rollouts using Fréchet Video Distance (FVD)~\cite{unterthiner2019fvd}, evaluated on three benchmarks: Waymo~\cite{sun2020scalability}, nuPlan, and nuPlan-turns~\cite{mousakhan2026overcoming}. nuPlan-turns is a subset of nuPlan dataset focused on turning scenarios. 800 sequences are used for nuPlan benchmark, and 400 sequences for Waymo and nuPlan-turns benchmarks.

\textbf{Long-horizon stability.}
In a predictive setting, standard FVD can be confounded by shared conditioning frames at the beginning of each rollout. Methods that prioritize perceptual fidelity or are optimized for long-chunk generation can achieve strong short-horizon FVD without truly capturing semantics or physical structure. 
To address this, we adopt the chunked-FVD metric from~\cite{mousakhan2026overcoming}, which measures FVD over consecutive 4-second windows. We additionally introduce \textbf{FVD-slope}, a scalar metric capturing the temporal degradation of video quality across consecutive chunks:
$\text{FVD-slope} = \frac{\text{FVD}_T - \text{FVD}_1}{T - 1}$, where $\text{FVD}_t$ denotes the FVD of the $t$-th chunk and $T$ is the total number of chunks. A lower FVD-slope indicates greater stability over long-horizon prediction.


\paragraph{Action controllability.} Given the importance of action-conditional generation for world models, we evaluate their ability to follow given steering commands in the form of 2D trajectories. Beyond the original trajectories, we evaluate on counterfactual trajectories with scaled speed and yaw rate (under- and oversteering), which cannot be inferred from the context frames and therefore test whether the model genuinely responds to the steering input.

\section{Results}~\label{sec:results}

\subsection{Comparison to State-of-the-art}

\paragraph{Long-horizon rollout performance}
The ability to perform long-horizon rollouts reflects the quality of the learned transition function -  how well the world model maps past states to future states. We first evaluate this by comparing driving world models using FVD. Table~\ref{table:main_results} (first three columns) reports FVD over the 6-second horizon window. 
Cosmos-v2.5 (version: Predict-2B, post-trained, Video2World) achieves the lowest FVD scores for nuPlan benchmarks. However, it is $2\times$ larger than our model and is trained on over $3\times$ more driving data from an undisclosed data mix. Despite this, our model outperforms Cosmos-v2.5 on the Waymo benchmark.
We further compare long-horizon rollout stability using chunked-FVD~(see Figure~\ref{fig:fvd_chunked}) and FVD-slope~(see Table~\ref{table:main_results}), where our model achieves the best scores, demonstrating significantly greater stability than all baselines including Cosmos-v2.5. Epona, which is trained only on nuPlan and nuScenes datasets, performs well on nuPlan benchmarks but fails to generalize to the unseen Waymo. 
Orbis shows the next-best rollout stability. It also uses two abstraction levels, but in a single branch setting and at same temporal resolution. A direct comparison with the Orbis (v1) model is added in Appendix~\ref{app:orbis_vs_hiera}. Please find qualitative examples of long-horizon prediction for different baselines on the project page.

\begin{table}[t]
    \centering
    \caption{FVD over 6s rollouts on nuPlan, Waymo and nuPlan-turns datasets. FVD-slope is shown for 4 chunks of 4 seconds each. Linear probing results are shown for semantic segmentation on Cityscapes and depth estimation on KITTI. *Epona is trained only on the nuPlan and nuScenes datasets. The training mix of Cosmos is not public.}
    \label{table:main_results}
    \vspace{2mm}
    \begin{tabular}{l|ccc|c|cc}
    \toprule
      & \multicolumn{3}{c}{FVD$\downarrow$} & FVD-slope$\downarrow$ & Segmentation & Depth\\
      Model  & nuPlan & Waymo & nuPlan-turns   & nuPlan-turns & mIoU$\uparrow$  & RMSE$\downarrow$\\
      \midrule
      Vista         & 289.95 & 351.42 & 353.27 & 434.67 & 39.76   & 4.884\\
       GEM          & 348.36 & 218.61 & 318.73  & 258.00 &   45.81   &  4.373  \\
      Orbis         & 132.25 & 180.54 & 239.05 & 99.67 &  52.39   &  4.321  \\
      \textcolor{gray}{Epona*} & \textcolor{gray}{251.81} & 373.03 & \textcolor{gray}{123.60} & \textcolor{gray}{93.89} & 35.81 & 5.720 \\
       Cosmos-v2.5*   & \textbf{67.74} & 116.12 & \textbf{126.26}  & 189.34 &  43.20    & 6.384 \\
      \midrule
      \textbf{Orbis 2} & 98.97  
      & \textbf{98.01} & 187.53 & \textbf{46.45} & \textbf{58.29}  & \textbf{4.157}\\
      \bottomrule
    \end{tabular}
\end{table}

\paragraph{Conditional trajectory evaluation}
To evaluate the responsiveness of our model to steering commands, we consider two settings. 1)~\textbf{Original trajectories}: using the unaltered ground truth steering signal corresponding to each validation sample (as done in~\cite{NEURIPS2024_a6a066fb,hassan2025gem,mousakhan2026overcoming}) and 2)~\textbf{Counterfactual trajectories}: obtained by altering the original odometry, to construct alternative steering events (e.g. under/oversteering). Specifically, we obtain four counterfactual trajectories per sample by scaling the original speeds and yaw rates by a factor 0.5 and 1.5. It is essential to evaluate steering in a counterfactual setting, because the original trajectories are partly leaked by the context frames, i.e. a model that cannot steer could still score well by relying on context only.
We evaluate the models by estimating the trajectory of the generated videos using an external inverse dynamics model, as done in~\cite{NEURIPS2024_a6a066fb,hassan2025gem,mousakhan2026overcoming}, specifically VGGT~\cite{wang2025vggt}. We then compute the average displacement error (ADE) between the estimated and ground truth trajectories.


\begin{wraptable}{r}{0.45\textwidth}
    \vspace{-\intextsep}
    \centering
    \small
    \setlength{\tabcolsep}{3pt}   
    \caption{Steering evaluation on original and counterfactual trajectories. ADE$\downarrow$ in meters.}
    \begin{tabular}{@{}lccccc@{}}
        \toprule
        & \multirow{2}{*}{Original} & \multicolumn{4}{c}{Counterfactual} \\
        \cmidrule(lr){3-6}
        & & \multicolumn{2}{c}{speed} & \multicolumn{2}{c}{yaw rate} \\
        \cmidrule(lr){3-4} \cmidrule(lr){5-6}
        Model & & $\times0.5$ & $\times1.5$ & $\times0.5$ & $\times1.5$ \\
        \midrule
        Orbis   & 2.30 & 1.39 & 3.27 & 1.31 & 3.69 \\
        Epona   & 1.40 & 1.76 & 2.10 & 1.81 & 2.42 \\
        \midrule
        \textbf{Orbis 2} & \textbf{1.24} & \textbf{0.22} & \textbf{1.70} & \textbf{0.59} & \textbf{2.38} \\
        \bottomrule
    \end{tabular}
    \label{tab:steering_complete}
    \vspace{-1em}
\end{wraptable}

Because VGGT trajectories are scale-ambiguous, we align each generated trajectory $\mathbf{x}$ to its real counterpart $\bar{\mathbf{x}}$ before computing ADE using the least-squares scale $s^*=\left(\sum_t \mathbf{x}_t^\top\bar{\mathbf{x}}_t\right)/\left(\sum_t \mathbf{x}_t^\top\mathbf{x}_t\right)$. This removes global scale error while preserving penalties for shape, direction, and curvature.
We evaluate one original and four counterfactual modes on nuPlan-turns with six-second drives. \mbox{\Cref{tab:steering_complete}} shows that our model is best on all cases. Qualitative examples are in Appendix~\mbox{\ref{app:counterfactual_steering}}.

\subsection{Strong predictor representations lead to stable world models}

We evaluate the internal representation quality of driving world models by linear probing on semantic segmentation and depth estimation (Table~\ref{table:main_results}). We observe a clear trend: models with stronger semantic and geometric understanding also achieve more stable long-horizon prediction, as measured by FVD-slope. 
Our model achieves the best results on both probing tasks and the long-horizon stability metric. Orbis (v1), which also learns strong representations, achieves the second-best long-horizon stability. Several baselines operate at significantly higher resolutions to maximize perceptual fidelity, yet our model is more stable at $512\times288$ resolution. 
This suggests that representation quality plays a complementary, and arguably more important role in world modeling than perceptual fidelity alone.

\subsection{Comparison between hierarchical variants}~\label{sec:hiera_designs}
\begin{figure}[h!]
\centering
\setlength{\tabcolsep}{1pt}
\begin{tabular}{c c c c c c}
 & \scriptsize 0.5s & \scriptsize 4s & \scriptsize 8s & \scriptsize 12s & \scriptsize 16s \\[2pt]
 \rotatebox{90}{\scriptsize Experiment-1} &
\includegraphics[width=0.16\textwidth]{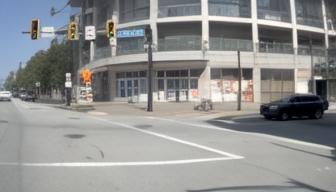} &
\includegraphics[width=0.16\textwidth]{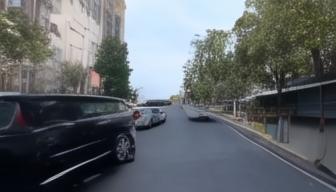} &
\includegraphics[width=0.16\textwidth]{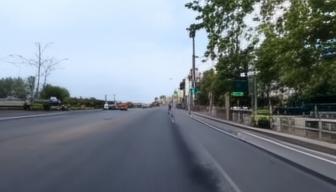} &
\includegraphics[width=0.16\textwidth]{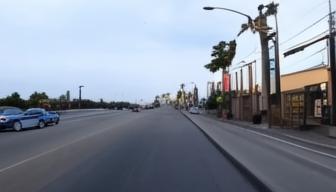} &
\includegraphics[width=0.16\textwidth]{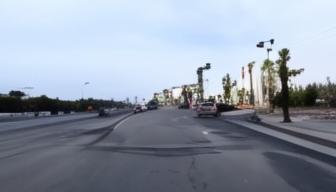} \\[2pt]
\rotatebox{90}{\scriptsize Experiment-2} &
\includegraphics[width=0.16\textwidth]{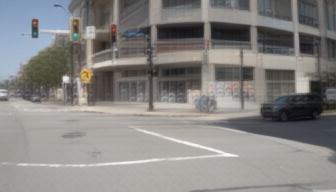} &
\includegraphics[width=0.16\textwidth]{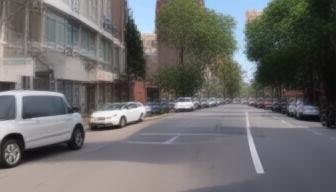} &
\includegraphics[width=0.16\textwidth]{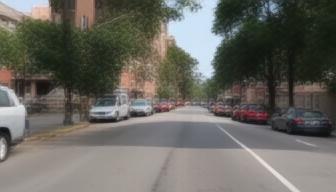} &
\includegraphics[width=0.16\textwidth]{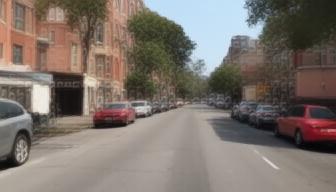} &
\includegraphics[width=0.16\textwidth]{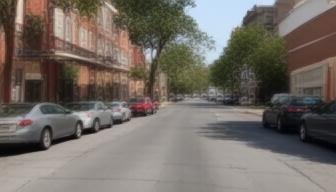} \\[2pt]
\rotatebox{90}{\scriptsize Exp-3 (ours)} &
\includegraphics[width=0.16\textwidth]{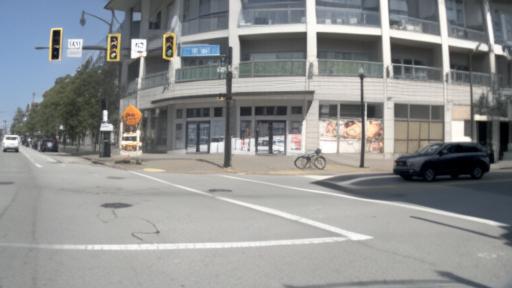} &
\includegraphics[width=0.16\textwidth]{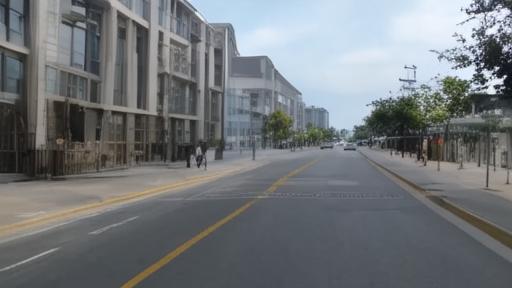} &
\includegraphics[width=0.16\textwidth]{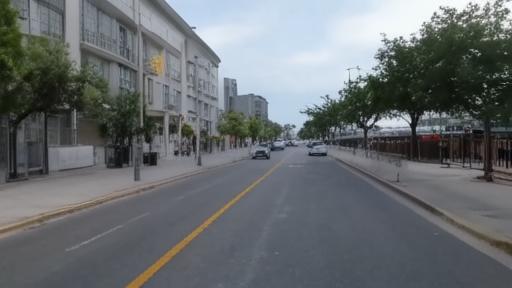} &
\includegraphics[width=0.16\textwidth]{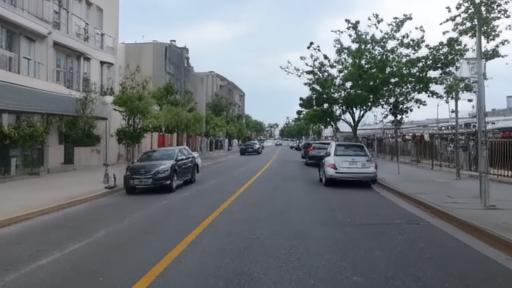} &
\includegraphics[width=0.16\textwidth]{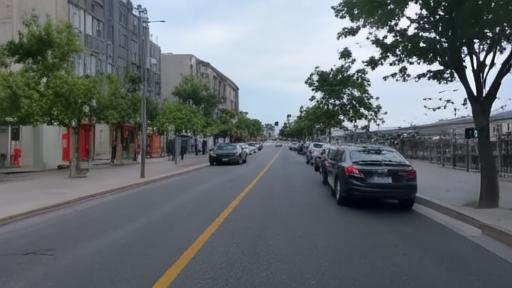} \\[2pt]
\rotatebox{90}{\scriptsize Experiment-4} &
\includegraphics[width=0.16\textwidth]{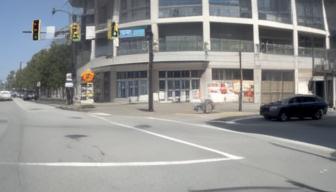} &
\includegraphics[width=0.16\textwidth]{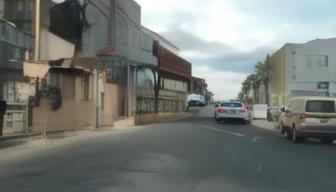} &
\includegraphics[width=0.16\textwidth]{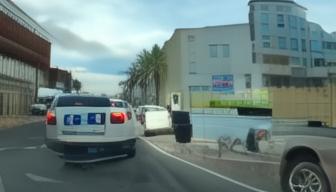} &
\includegraphics[width=0.16\textwidth]{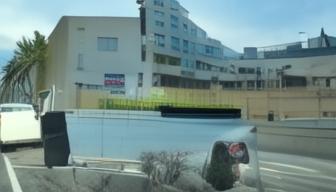} &
\includegraphics[width=0.16\textwidth]{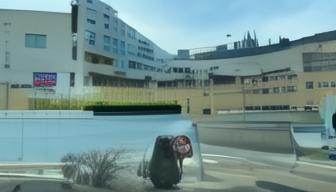} \\[2pt]
\rotatebox{90}{\scriptsize Experiment-5} &
\includegraphics[width=0.16\textwidth]{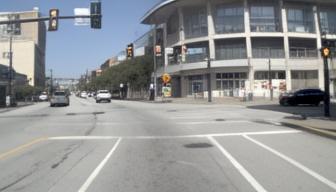} &
\includegraphics[width=0.16\textwidth]{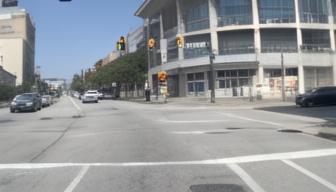} &
\includegraphics[width=0.16\textwidth]{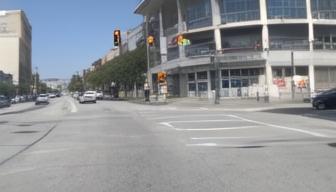} &
\includegraphics[width=0.16\textwidth]{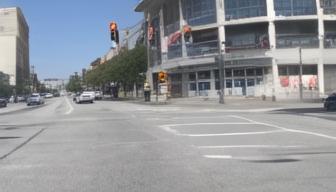} &
\includegraphics[width=0.16\textwidth]{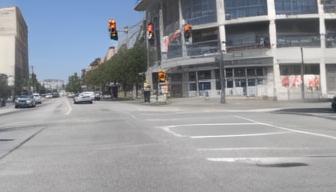} \\[2pt]
\end{tabular}
\caption{Qualitative examples showing behavior of different hierarchy variants discussed in Table~\ref{tab:hierarchy_ablation}. We observe that Exp-2 rollouts are stable but lacks details, Exp-4 drifts out-of-distribution and Exp-5 suffers from loss in motion.}
\label{fig:grid}
\end{figure}

\Cref{tab:hierarchy_ablation} compares the design choices for the top and bottom predictors (illustrated in~\cref{fig:framework}), varying the level of latent abstraction, frame rate, and training mode. 
The first two rows establish flat (single-branch) baselines at different abstraction levels.
The detail-level model (Exp.~1) achieves strong but progressively degrading FVD and learns weak internal representations, whereas the abstract-level model (Exp.~2) yields substantially richer representations and remarkably stable rollout quality over long horizons, despite its high absolute FVD.
The hierarchical configurations (Exps.~3–5) all fix the bottom predictor to operate on detailed latents at 10~fps, and vary the top predictor.
Exp.~3 shows the best hierarchical design, where the top predictor operates on high abstraction latents  at a longer horizon of 2 fps: it achieves both strong representations and the best FVD-slope with a competitive absolute FVD performance.
Exp.~4 has high FVD-slope: the subpar long-horizon rollouts of the top predictor in the detailed latent space cause the model to drift out-of-distribution.
In Exp.~5, the top predictor operating in abstract space in high-temporal resolution leads to stopping behavior in rollouts. We attribute this to the abstract latent space is not detailed enough to capture small motion at 10~fps, which leads to loss in motion signal. All bottom predictors do not learn strong representations due to their detailed predictive task with strong conditioning.
The best design is hierarchical with an abstract-latent top predictor at 2~fps (Exp.~3): it best balances overall (FVD) and long-horizon video fidelity (FVD-slope), with the strongest representations. Figure~\ref{fig:grid} qualitatively confirms this: Exp.~1 drifts, Exp.~2 is stable but lacks detail, Exp.~4 diverges out-of-distribution, and Exp.~5 progressively loses motion, whereas Exp.~3 remains stable and perceptually detailed.

\begin{table*}[h]
    \centering
    \small
     \caption{Hierarchy design study across abstractions and temporal resolutions. All experiments have comparable parameter counts. Exp.~3 is our final hierarchical setting. Top/Bottom refers to predictor positions in~\cref{fig:framework}. FVD and FVD-slope are reported on the nuPlan-turns benchmark.}
      \begin{tabular}{llcccccc}
      \toprule
      \multirow{2}{*}{Exp.} & Model    & Abstraction  & Frame-rate & Sem. Seg.     & Depth    & FVD$\downarrow$  & FVD-slope$\downarrow$ \\
      & branch  & detail/abstract        & fps      &  mIoU $\uparrow$   & RMSE $\downarrow$ & 16s & 4 x 4s\\
      \midrule
      1. & Single-branch       & detail  & 10 &   32.72    &  6.107      & 170.02 & 101.63 \\ 
      \midrule
      2. & Single-branch         & abstract  & 10   & 47.88  &  4.624       & 490.04    & 48.38 \\
      \midrule
    \multirow{2}{*}{3.} & Top    &  abstract    & 2    & 49.39   &  4.498  & \multirow{2}{*}{209.15} & \multirow{2}{*}{47.87} \\ %
     & {} \Ldownarrow {} {} Bottom  & detail      & 10     & 31.97    &  6.396   &  & \\ 
    
     \midrule
     \multirow{2}{*}{4.} & Top  &  detail     & 2     & 41.02   &  5.256  & \multirow{2}{*}{207.71} & \multirow{2}{*}{81.40} \\ %
      & {} \Ldownarrow {} {} Bottom  &  detail   & 10    &  25.52    & 6.514      &  & \\ 
         \midrule 
    \multirow{2}{*}{5.} & Top     &  abstract    & 10     & 49.27     &  4.460  & \multirow{2}{*}{551.38}  & \multirow{2}{*}{137.07} \\ 
     & {} \Ldownarrow {} {} Bottom   & detail      & 10     & 31.46  &  6.311    &  &  \\ 
    \bottomrule
    \end{tabular}
    \label{tab:hierarchy_ablation}
\end{table*}

\subsection{Training strategy: diffusion forcing + teacher forcing}\label{sec:df_tf}

We evaluate the effect of diffusion forcing pre-training followed by teacher forcing fine-tuning on the representation quality of the abstract predictor. Table~\ref{tab:df_tf} reports linear probing results for semantic segmentation and depth estimation. Diffusion forcing improves over teacher forcing alone on both tasks, and subsequent teacher forcing fine-tuning provides additional gains, yielding the best overall probing performance among our predictor variants.
The final predictor outperforms the abstract tokenizer based on DINOv2-B on both probing tasks, improving mIoU from $57.19$ to $58.29$ and reducing RMSE from $4.574$ to $4.157$. 
The same trend holds for the predictor trained on pixel reconstruction-based tokenizer latents: diffusion forcing improves representation quality at every block of the predictor, on both semantic segmentation and depth estimation, rather than only at the output. The per-block probing curves are reported in Figure~\mbox{\ref{fig:linear_probing}} in the Appendix. 

DINOv2-Base shows stronger overall performance in segmentation probing, but our predictor achieves better performance on frequent driving classes such as road, car, sidewalk, and vegetation, while underperforming on rarer classes such as bus, train, and motorcycle (see Appendix~\ref{sec:classwise_miou} for class-wise performance). Compressing DINOv2 features into the tokenizer reduces depth estimation performance, but our training strategy significantly narrows the gap to DINOv2. With an identical DPT~\cite{ranftl2021vision} decoder trained on frozen features (Table~\ref{tab:kitti_dpt_probe}), Orbis~2 slightly outperforms DINOv2-B and V-JEPA~2.1~\cite{mur2026v} on single frames. The gap widens with temporal context, suggesting that the predictor's features integrate geometric cues across frames.



\begin{table*}[h]
    \centering
    \small
    \begin{minipage}[t]{0.55\textwidth}
        \centering
        \captionof{table}{Ablation of the proposed training strategy. All metrics are reported for the abstract branch including tokenizer and predictor. Full-scale setting used.}
        \label{tab:df_tf}
        \vspace{2mm}
        \begin{tabular}{lcc}
        \toprule
       \multirow{2}{*}{Model}       & Segmentation & Depth \\
            &  mIoU $\uparrow$ (val loss$\downarrow$) & RMSE $\downarrow$ \\
        \midrule
        Abstract Tokenizer                 &   57.19  (0.265)     &  4.574    \\
        \midrule
        Teacher forcing~(TF) only          &   55.01  (0.270)     &  4.355   \\
        Diffusion forcing~(DF) only        &   58.03   (0.254)    &  4.274  \\
        DF + TF (Orbis 2)               &   58.29   (0.250)    &  4.157  \\
        \midrule
        DINOv2-Base   & 60.01 (0.284)   & 3.890   \\
        \bottomrule
        \end{tabular}
    \end{minipage}%
    \hfill
    \begin{minipage}[t]{0.40\textwidth}
        \centering
        \small
        \captionof{table}{Inference speed and VRAM requirements across models, measured on a H100 GPU.}
        \label{tab:inf_speed}
        \begin{tabular}{l|c|c}
        \toprule
        Method  & FPS$\uparrow$ & VRAM $\downarrow$\\
        \midrule
        Cosmos v2.5  & 0.66 & 22  \\
        Vista   & 0.58 & 86  \\
        GEM     & 0.44 & 45  \\
        Epona   & 0.4 & 17  \\
        Orbis   & 0.70 & 24  \\
        \midrule
        \textbf{Orbis 2}    &  4.37    &  19   \\ 
        \textbf{Orbis 2 Distilled}    &  17.40    &  19   \\
        \bottomrule
        \end{tabular}
    \end{minipage}
\end{table*}

\begin{table}[t]
\centering
\small
\caption{Monocular depth on a drive-disjoint KITTI split from frozen backbones, each with an identical, freshly trained DPT decoder. Six frames: target plus five past frames at 2\,Hz. DINOv2 is evaluated on single frames only. RMSE in metres; best per block in bold. Full protocol and further baselines in Appendix~\ref{app:linear_probe_details}.}
\label{tab:kitti_dpt_probe}
\setlength{\tabcolsep}{4pt}
\begin{tabular}{lrcccccc}
\toprule
 & & \multicolumn{3}{c}{Single frame} & \multicolumn{3}{c}{Six frames @ 2\,Hz} \\
\cmidrule(lr){3-5}\cmidrule(lr){6-8}
Backbone & Params & AbsRel$\downarrow$ & RMSE$\downarrow$ & $\delta_1\uparrow$ & AbsRel$\downarrow$ & RMSE$\downarrow$ & $\delta_1\uparrow$ \\
\midrule
DINOv2-B         & 87M   & 0.067 & 3.035 & 0.950 & --    & --    & --    \\
V-JEPA 2.1 ViT-g & 1013M & 0.066 & 3.043 & 0.951 & 0.069 & 3.067 & 0.949 \\
Orbis 2 (Abstract pred. only)     & 599M  & \textbf{0.065} & \textbf{3.016} & \textbf{0.953} & \textbf{0.060} & \textbf{2.791} & \textbf{0.966} \\
\bottomrule
\end{tabular}
\end{table}

\begin{wraptable}{r}{0.6\textwidth}
    \centering
    \small
    \setlength{\tabcolsep}{2.5pt}
    \renewcommand{\arraystretch}{0.8}
    \caption{Isolating the training strategy from the hierarchical architecture. Top/Bottom refers to position of predictors in Table~\ref{fig:framework}. DF+TF improves representation quality, consistent with Table~\ref{tab:df_tf}, and substantially improves long-horizon stability. FVD and FVD-slope are reported on the nuPlan-turns.}
    \begin{tabular}{@{}lccccc@{}}
        \toprule
        Model & Training & Sem. Seg. & Depth & FVD$\downarrow$ & FVD-slope$\downarrow$ \\
        branch & mode & mIoU$\uparrow$ & RMSE$\downarrow$ & 16s & 4 x 4s \\
        \midrule
        Top & TF
            & \multirow{2}{*}{48.01}
            & \multirow{2}{*}{4.606}
            & \multirow{2}{*}{239.46}
            & \multirow{2}{*}{107.21} \\
        \Ldownarrow Bottom & TF & & & & \\
        \midrule
        Top & DF+TF
            & \multirow{2}{*}{\textbf{49.39}}
            & \multirow{2}{*}{\textbf{4.490}}
            & \multirow{2}{*}{\textbf{209.15}}
            & \multirow{2}{*}{\textbf{47.87}} \\
        \Ldownarrow Bottom & TF & & & & \\
        \bottomrule
    \end{tabular}
    \label{tab:training_mode_ablation}
\end{wraptable}

\textbf{Ablation study}
To isolate the contribution of the hierarchical architecture from the proposed training strategy, we conduct an ablation comparing two hierarchical setups with identical model size, compute, and data, differing only in how the abstract predictor is trained: one with teacher forcing (TF) only, and one with our proposed diffusion-forcing + teacher-forcing (DF+TF) strategy (the bottom/detail predictor uses teacher forcing in both). Results on representation quality follow the same trend as Table~\ref{tab:df_tf}, where DF+TF performs better than TF-only. The long-horizon stability with TF-only training is significantly worse compared to DF+TF training. The TF-only hierarchical model produces FVD-slope 107.21, compared to FVD-slope 47.87 with DF+TF training. Results are summarized in Table~\ref{tab:training_mode_ablation}.



\subsection{More analysis}

\paragraph{Model efficiency.}

Our final 1.07B-parameter model is trained in under 8k H100-GPU-hours, substantially less compute than comparable driving world models. GAIA-1's 6.5B world model required ~23k A100-hours, Epona used ~16k A100-hours for a 2.5B model, and Orbis used ~8k A100-hours to train a 469M-parameter model, less than half our size. Our model achieves the fastest inference at 4.37 FPS while requiring only 19\,GB VRAM, at least $5×\times$ faster than all baselines, shown in Table~\ref{tab:inf_speed}.  
Further, we apply teacher-forced causal consistency distillation following Causal Forcing++~\cite{zhao2026causal}, building on consistency distillation~\cite{song2023consistency}. Using consistency distillation for the abstract predictor and heun sampler~\cite{karras2022elucidating,heun1900neue} for the detail predictor, inference latency improves from 4.37 to 17.40 FPS with marginal drop in performance.



\textbf{Abstract latent design.} Two choices define our abstract latent space: whether to compress DINO at all, and how far. A natural alternative is to predict directly in the high-dimensional DINO space, as proposed in Representation Autoencoders (RAE)~\mbox{\cite{zheng2025diffusion}}, which could in principle replace our compressed latent with raw DINO features; and given compression, the latent width trades semantic capacity against how easily flow matching can model the space. We find that predicting in full DINO space is prohibitive for autoregressive rollouts, with frame quality degrading progressively over the horizon, whereas our compressed latent retains the semantic structure of DINO at a dimensionality that flow matching models stably. Within the compressed space, representation quality improves monotonically with width, but long-horizon stability is best at 16 dimensions and degrades on either side of it, so 16-dim gives the best overall trade-off. The qualitative comparison and the dimensionality study supporting both conclusions are reported in Appendix~\mbox{\ref{app:dino_compressed}} and Appendix~\mbox{\ref{app:latent_dim}}.

\section{Conclusion}~\label{sec:conclusion}
We demonstrated that the proposed hierarchical approach makes effective use of abstract and detailed latent spaces, yielding a world model that combines high-fidelity short-term predictions with  stability over long horizons. The model is steerable through action conditioning. We further demonstrated a strong correlation between representation quality and rollout stability, and showed that a diffusion forcing pre-training strategy consistently strengthens the learned representations.

\textbf{Limitations.} Although diffusion forcing improves predictor representations, the reason for this improvement is not fully studied in this work. We hypothesize that independent noise levels across frames act as implicit data augmentation, while different noise patterns encourage the model to learn from related tasks such as denoising, interpolation, and forecasting. We do not study these factors separately in this work, and leave a more detailed analysis of noise schedules and corruption patterns to future work. Our hierarchy relies on separate abstract and fine-grained latent spaces, each optimized with a different objective. While this separation clarifies the role of each level, it limits the interaction between abstraction levels. Improving the connection between the two spaces is an important direction for future work. 

\textbf{Societal Impact.} 
At this stage, the work remains an early research contribution, and any practical societal effects will depend on substantial future progress in model design, scaling, validation, and system integration.

\section*{Acknowledgment}
This work was funded by the German Federal Ministry for Economic Affairs and Energy within the project “NXT GEN AI METHODS" (19A23014R). 
The authors gratefully acknowledge the Gauss Centre for Supercomputing e.V. (www.gauss-centre.eu) for providing computing time through the John von Neumann Institute for Computing (NIC) on the GCS Supercomputer JUWELS~\cite{JUWELS} and JUPITER at Jülich Supercomputing Centre (JSC). This project was selected as a winner of Gauss AI Compute Competition, organized by the Gauss Centre for Supercomputing (GCS) and Jülich Supercomputing Centre (JSC).
The project was accomplished under GCS compute grants - \textit{multiscale-wm} and \textit{nxtaim-1}.
Further local compute resources were funded through the Deutsche Forschungsgemeinschaft (DFG, German Research Foundation) under grant numbers 417962828, 539134284, through EFRE (FEIH\_2698644), and the state of Baden-Württemberg. 
\newsavebox{\logostack}
\begin{center}
\sbox{\logostack}{%
  \begin{minipage}[c]{0.35\textwidth}
    \centering
    \includegraphics[width=\linewidth]{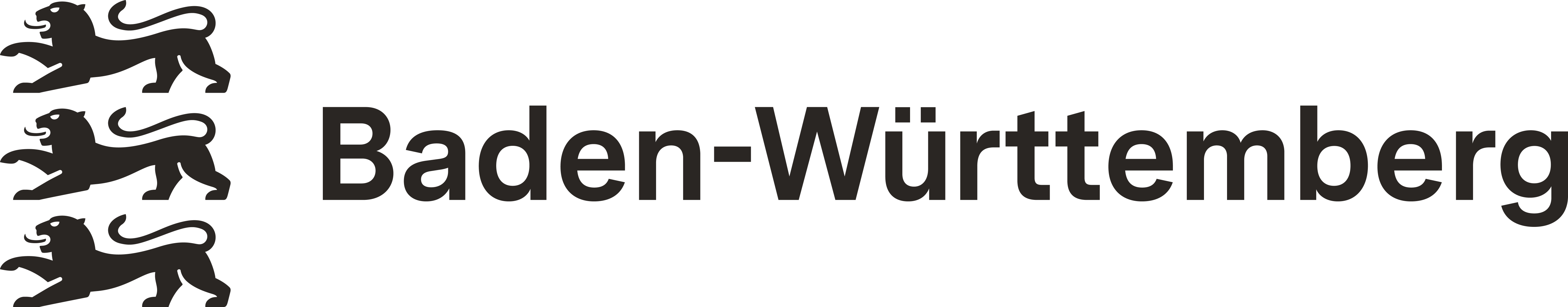}\\[2ex]
    \includegraphics[width=\linewidth]{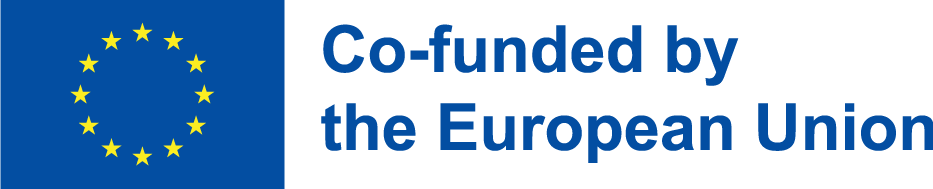}
  \end{minipage}}%
\usebox{\logostack}\hspace{2em}%
$\vcenter{\hbox{\includegraphics[height=1.3\dimexpr\ht\logostack+\dp\logostack\relax]{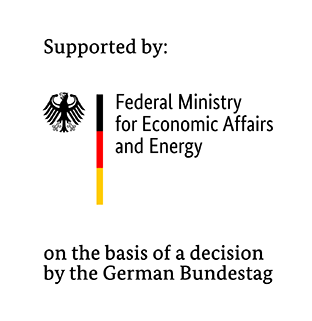}}}$
\end{center}
We thank NATIX for providing access to their large-scale, crowd-sourced, multi-camera driving dataset~\cite{natix2026_multi_camera_driving_dataset} in support of this open-source research. We thank Marcel Aach for his technical support and assistance with the GCS HPC cluster. We also thank Muhammad Ali for improving the model latency for the Hugging Face demo.


\bibliographystyle{plain}


\newpage
\appendix
\section{Appendix}

\subsection{Class-wise semantic segmentation probing results}~\label{sec:classwise_miou}
In Table~\ref{tab:iou_comparison}, we compare class-wise semantic segmentation probing performance for the tokenizer, predictor, and DINOv2-Base representations. DINOv2-Base achieves stronger overall segmentation performance, but our predictor performs better on frequent driving classes such as road, car, sidewalk, and vegetation, improving by $2$--$5$ mIoU points. In contrast, it underperforms on rarer classes such as bus, train, rider, and motorcycle, with a gap of $5$--$20$ mIoU points.

\begin{table}[h!]
\centering
\caption{Comparison of our tokenizer and abstract predictor model with DINOv2-base. Per-class IoU for semantic segmentation reported on Cityscapes. $\Delta$ columns show the predictor's change relative to our tokenizer and to DINOv2-B.}
\label{tab:iou_comparison}
\begin{tabular}{lccccc}
\toprule
\multirow{2}{*}{Class} & \multirow{2}{*}{Dino v2-B} & Ours & Ours & $\Delta$ Predictor & $\Delta$ Predictor \\
 & & tokenizer & predictor & $-$ Tokenizer & $-$ Dino v2-B \\
\midrule
Road          & 94.88 & 95.3 & 95.9 & \textcolor{green!60!black}{+0.6} & \textcolor{green!60!black}{+1.02} \\
Sidewalk      & 66.58 & 67.6 & 71.0 & \textcolor{green!60!black}{+3.4} & \textcolor{green!60!black}{+4.42} \\
Building      & 82.98 & 84.4 & 84.8 & \textcolor{green!60!black}{+0.4} & \textcolor{green!60!black}{+1.82} \\
Wall          & 48.28 & 47.7 & 47.9 & \textcolor{green!60!black}{+0.2} & \textcolor{red}{$-$0.38} \\
Fence         & 45.84 & 44.8 & 44.9 & \textcolor{green!60!black}{+0.1} & \textcolor{red}{$-$0.94} \\
Pole          & 17.54 & 16.7 & 18.5 & \textcolor{green!60!black}{+1.8} & \textcolor{green!60!black}{+0.96} \\
Traffic Light & 33.97 & 30.1 & 31.5 & \textcolor{green!60!black}{+1.4} & \textcolor{red}{$-$2.47} \\
Traffic Sign  & 42.93 & 41.4 & 43.6 & \textcolor{green!60!black}{+2.2} & \textcolor{green!60!black}{+0.67} \\
Vegetation    & 80.87 & 83.3 & 83.6 & \textcolor{green!60!black}{+0.3} & \textcolor{green!60!black}{+2.73} \\
Terrain       & 53.05 & 53.5 & 53.4 & \textcolor{red}{$-$0.1} & \textcolor{green!60!black}{+0.35} \\
Sky           & 82.11 & 86.6 & 87.0 & \textcolor{green!60!black}{+0.4} & \textcolor{green!60!black}{+4.89} \\
Person        & 54.18 & 53.0 & 52.7 & \textcolor{red}{$-$0.3} & \textcolor{red}{$-$1.48} \\
Rider         & 34.04 & 28.1 & 28.2 & \textcolor{green!60!black}{+0.1} & \textcolor{red}{$-$5.84} \\
Car           & 84.27 & 85.0 & 86.4 & \textcolor{green!60!black}{+1.4} & \textcolor{green!60!black}{+2.13} \\
Truck         & 70.25 & 67.0 & 67.1 & \textcolor{green!60!black}{+0.1} & \textcolor{red}{$-$3.15} \\
Bus           & 77.80 & 63.4 & 69.5 & \textcolor{green!60!black}{+6.1} & \textcolor{red}{$-$8.30} \\
Train         & 70.57 & 48.5 & 50.0 & \textcolor{green!60!black}{+1.5} & \textcolor{red}{$-$20.57} \\
Motorcycle    & 45.48 & 37.4 & 38.1 & \textcolor{green!60!black}{+0.7} & \textcolor{red}{$-$7.38} \\
Bicycle       & 54.57 & 52.9 & 53.1 & \textcolor{green!60!black}{+0.2} & \textcolor{red}{$-$1.47} \\
\midrule
mIoU          & 60.01 & 57.2 & 58.3 & \textcolor{green!60!black}{+1.1} & \textcolor{red}{$-$1.71} \\
\bottomrule
\end{tabular}
\end{table}

\subsection{Single-branch, mixed abstraction}~\label{app:orbis_vs_hiera}
Table~\ref{tab:imagefolder} shows the comparison between our model and a single branch model with mixed abstraction levels, combined via token factorization~\cite{li2024imagefolder}, as in~\cite{mousakhan2026overcoming}. Despite good FVD and representation quality, video fidelity degrades faster over time, as shown by the higher FVD-slope.

\begin{table*}[!h]
    \centering
    \caption{Comparison of single-branch and hierarchical predictors.}
    \begin{tabular}{lccccc}
        \toprule
        & Parameters & Sem. Seg. & Depth & FVD$\downarrow$ & FVD-slope$\downarrow$ \\
        Model & M & mIoU $\uparrow$ & RMSE $\downarrow$ & 16s & 4 $\times$ 4s \\
        \midrule
        Single-branch, mixed & 512 & 47.16 & 4.680 & 184.69 & 96.55 \\
        Hierarchical (ours) & 257+279 & 49.39 & 4.498 & 209.15 & 47.87 \\
        \bottomrule
    \end{tabular}
    \label{tab:imagefolder}
\end{table*}

\subsection{Qualitative rollouts with full DINO space vs compressed DINO space}~\label{app:dino_compressed}


RAE~\mbox{\cite{zheng2025diffusion}}~could in principle serve as the abstract predictor in our hierarchy by replacing our compressed latent with raw DINO features. However, its dimensionality is prohibitive for autoregressive rollouts, frame quality degrades progressively over the prediction horizon when the predictor operates directly in full DINO space. Our compressed DINO latent retains the semantic structure of DINO at reduced dimensionality, which proves essential for stable long-horizon prediction with a flow matching objective.

\begin{figure}[h!]
    \centering
    \includegraphics[width=\linewidth]{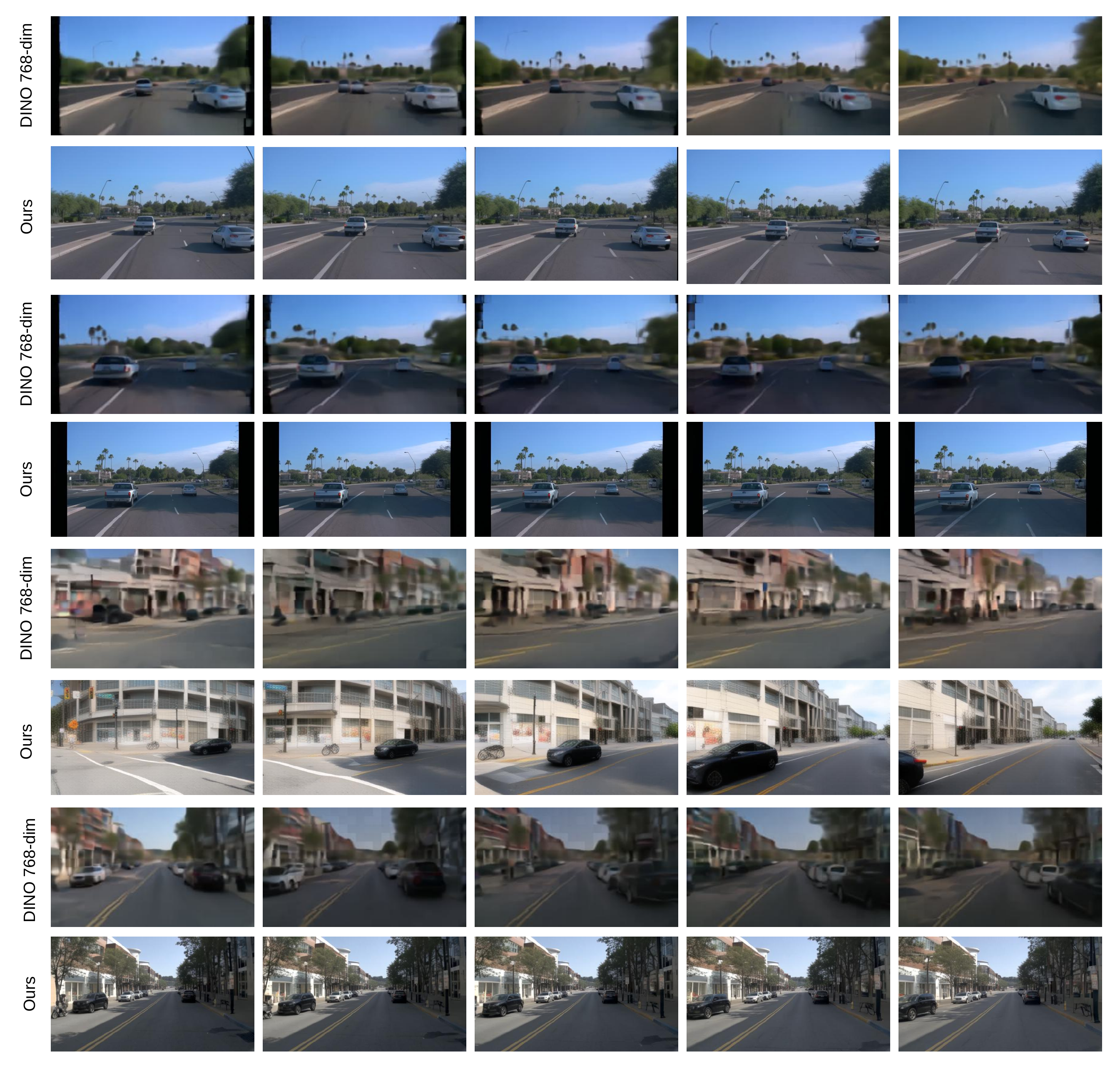}
    \caption{Comparison of raw DINO features versus our compressed DINO latent as the tokenizer for the abstract predictor. (Rollouts at 2hz)}
    \label{fig:more_rae}
\end{figure}

\Cref{fig:more_rae} shows qualitative results comparing 1-second rollouts generated using our compressed DINO latent targets versus the full 768-dim DINOv2 latent targets. Rollouts in the full-dimensional DINO space degrade rapidly, while the compressed DINO latent space is effectively modeled by the flow-matching objective, producing stable predictions throughout the horizon.

\subsection{Qualitative examples for different hierarchy variants}~\label{app:hiera_variants}
In \Cref{fig:grid}, we present qualitative examples illustrating the failure modes of the hierarchical design variants discussed in Section~\ref{sec:hiera_designs}. Experiment-1, which operates in single-branch mode at the detailed abstraction level, produces rollouts that drift away from the starting distribution. Experiment-2, a single-branch model at the abstract level, yields stable rollouts but fails to capture fine-grained motion and perceptual detail. Experiment-3 corresponds to our proposed hierarchical design, achieving stable rollouts with high perceptual quality. Experiment-4, which uses detailed latents for the top branch, diverges out-of-distribution rapidly, resulting in poor rollouts. Experiment-5, which operates abstract latents at high temporal frequency, fails to capture detailed motion, leading to motion degradation despite retaining high perceptual quality.

\subsection{Choice of abstract latent dimensionality}~\label{app:latent_dim}
Our choice of 16-dim for the abstract latent was initially guided by prior work (Orbis~\mbox{\cite{mousakhan2026overcoming}}), which uses a 16-dim DINO-aligned latent. To assess how sensitive performance is to this choice in our setup, we tested 8, 16 and 32-dim abstract latents at reduced scale, training the abstract predictor with our proposed strategy (DF+TF) and evaluating representation quality and FVD-slope over 32-second rollouts (4 chunks of 8s) at 2 fps. Results are summarized in Table~\mbox{\ref{tab:dino_dim}}.
Representation quality improves monotonically with dimensionality from 8 to 32, but long-horizon stability measured with FVD-slope is best at 16-dim, degrading at both 8 and 32. The dimensionality therefore offers a trade-off between representation quality and long-horizon stability, with 16-dim giving the best stability.

\begin{table}[!htb]
    \centering
    \small
    \caption{\textbf{Effect of abstract latent dimensionality} (DF+TF, reduced scale; FVD-slope over 32\,s at 2\,fps). Representation quality increases with dimensionality; stability is best at 16 dims.}
    \begin{tabular}{cccc}
    \toprule
    compressed-DINOv2   & Semantic Seg.        & Depth            & FVD-slope$\downarrow$ \\
    dimensionality    & mIoU $\uparrow$  & RMSE $\downarrow$ & 4 x 8s \\
    \midrule
    8  & 48.42 & 4.547 & 66.54 \\
    16 & 49.39 & 4.498 & \textbf{50.71} \\
    32 & \textbf{50.52} & \textbf{4.418} & 62.49 \\
    \bottomrule
    \end{tabular}
    \label{tab:dino_dim}
\end{table}

\subsection{Counterfactual steering}~\label{app:counterfactual_steering}
\paragraph{Evaluation details}
Evaluating with counterfactual steering inputs comes with additional challenges compared to the standard real-only steering evaluation done in~\cite{NEURIPS2024_a6a066fb,hassan2025gem,mousakhan2026overcoming}, where the inverse dynamics model is applied to real and generated videos alike, compensating for scale ambiguities. Here, we compare the estimated generated trajectories with the actual input trajectories, but this exposes us to scale mismatches due to the fact that the output of VGGT is not in metric space. As a solution, for each pair of real and generated trajectories $(\bar{\mathbf{x}}, \mathbf{x})$ we compute an optimal scale parameter for the latter:
\begin{equation}
s^* =
  \frac{
  \sum_{t=1}^{T} \mathbf{x}_t^\top \bar{\mathbf{x}}_t
  }{
  \sum_{t=1}^{T} \mathbf{x}_t^\top \mathbf{x}_t
  }
\end{equation}
and rescale it accordingly before computing ADE. This still penalizes shape, direction, and curvature, but it largely removes global trajectory-scale error.

\begin{figure}[ht]
\centering
\setlength{\tabcolsep}{1pt}
\renewcommand{\arraystretch}{1.1}

\begin{tabular}{c c c c c c c}
\rotatebox{90}{\tiny \hspace{0.7em} Original} &
\includegraphics[width=0.15\textwidth]{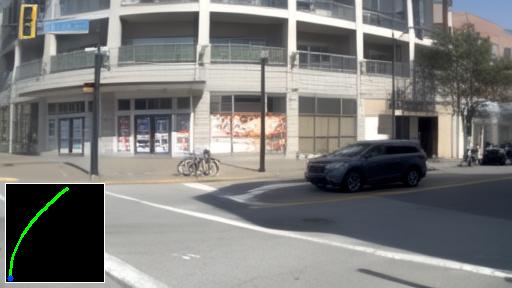} &
\includegraphics[width=0.15\textwidth]{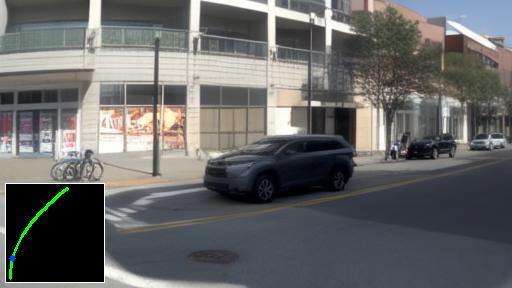} &
\includegraphics[width=0.15\textwidth]{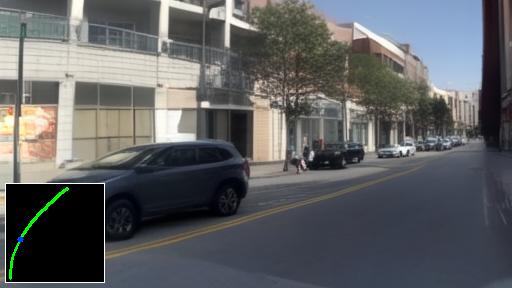} &
\includegraphics[width=0.15\textwidth]{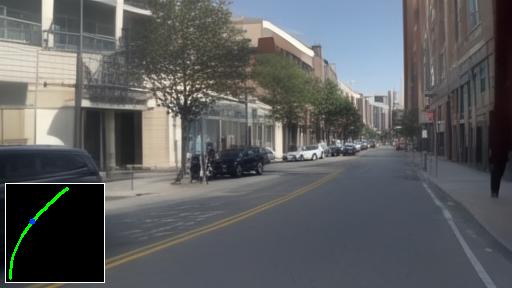} & 
\includegraphics[width=0.15\textwidth]{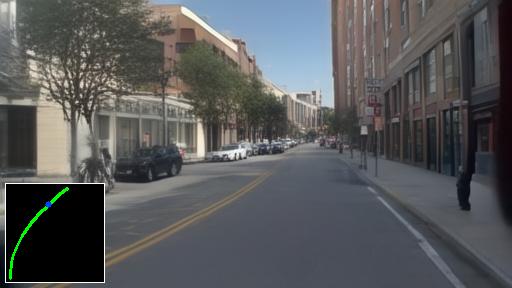} & 
\includegraphics[width=0.15\textwidth]{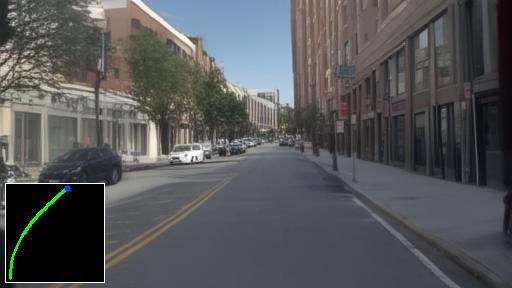} 
\\
\rotatebox{90}{\tiny \hspace{1.5em}$\times0.5$} &
\includegraphics[width=0.15\textwidth]{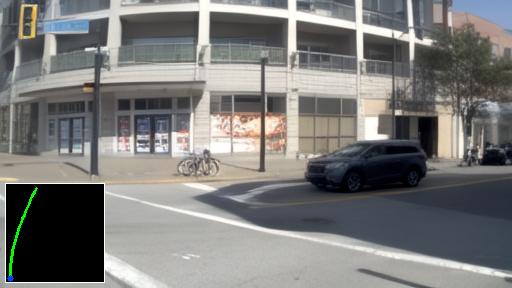} &
\includegraphics[width=0.15\textwidth]{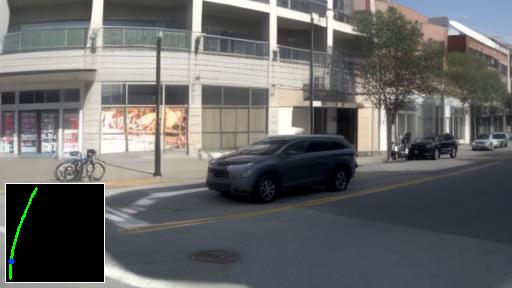} &
\includegraphics[width=0.15\textwidth]{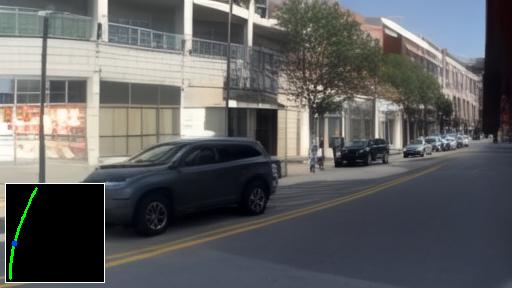} &
\includegraphics[width=0.15\textwidth]{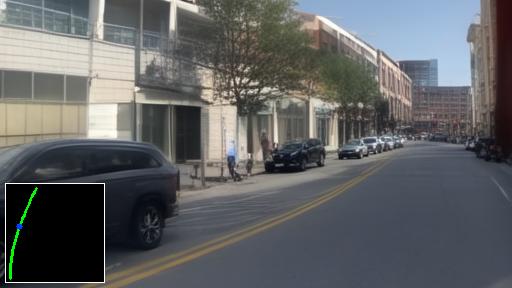} & 
\includegraphics[width=0.15\textwidth]{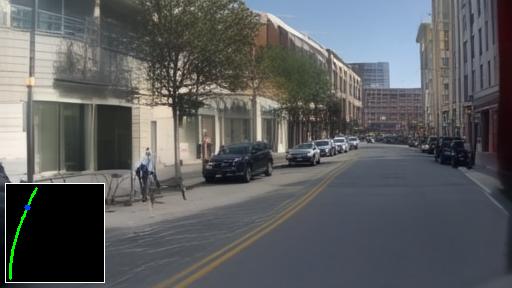} & 
\includegraphics[width=0.15\textwidth]{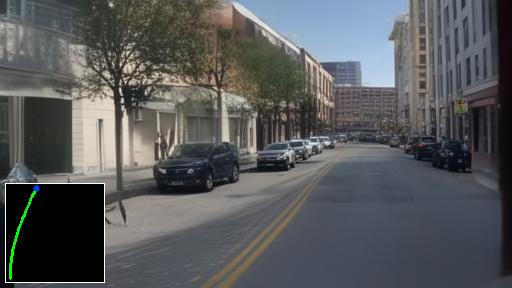} 
\\
\rotatebox{90}{\tiny \hspace{1.5em}$\times1.5$} &
\includegraphics[width=0.15\textwidth]{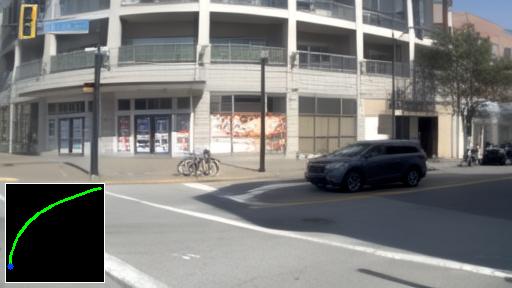} &
\includegraphics[width=0.15\textwidth]{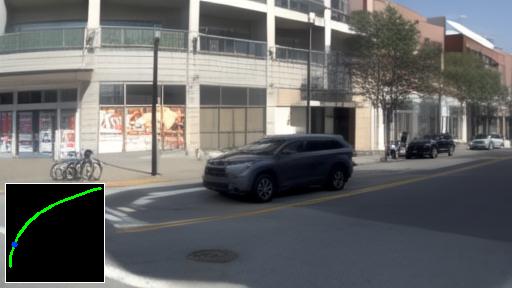} &
\includegraphics[width=0.15\textwidth]{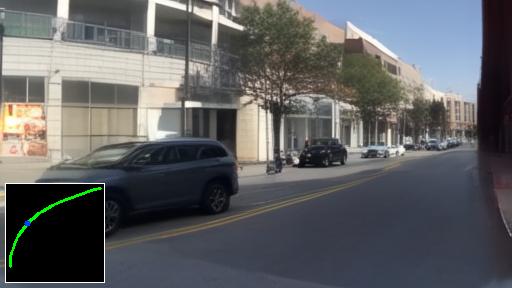} &
\includegraphics[width=0.15\textwidth]{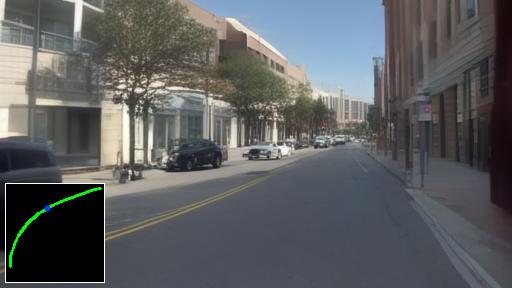} & 
\includegraphics[width=0.15\textwidth]{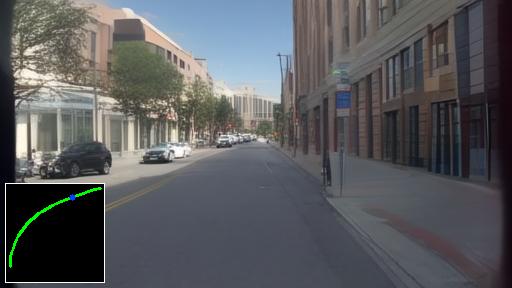} & 
\includegraphics[width=0.15\textwidth]{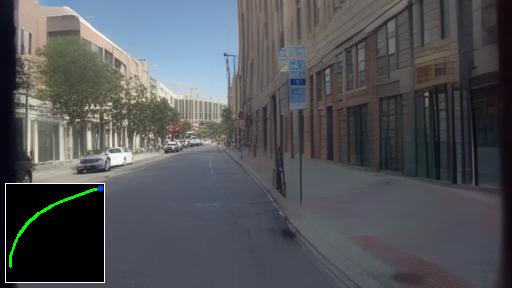} 
\\
& \scriptsize 0s & \scriptsize 0.5s & \scriptsize 1s & \scriptsize 1.5s & \scriptsize 2s & \scriptsize 2.5s \\
\end{tabular}

\caption{Example of counterfactual steering. Top row: the original steering trajectory, the model performs the turn as intended. Second row: the yaw rate was scaled by 0.5 (understeering), the model follows the command and executes a more gentle turn. Third row: the yaw rate was scaled by 1.5 (oversteering), the model follows the command and executes a sharper turn.}
\label{fig:steering_examples}

\end{figure}

Figure~\ref{fig:steering_examples} shows an example of steering the model on counterfactual trajectories. It is important for a world model to be able to follow commands that deviate from the original context-bound trajectories, even if they lead to out-of-distribution or illegal maneuvers.

\subsection{Model details}~\label{app:model_details}
\label{app:tokenizer}

\textbf{Tokenizer.} We train two hybrid tokenizers with separate encoder and decoder branches for detail and abstract latents. The hybrid architecture follows the design proposed by Mousakhan et al.~\cite{mousakhan2026overcoming}. Both encoders are initialized from a MAE-pretrained~\cite{he2022masked} ViT with patch size $16 \times 16$, producing $768$-dimensional embeddings that are quantized into $16$-dimensional codes.

\textbf{Abstract long-horizon predictor.}
This predictor generates semantic latents at 2\,Hz autoregressively with \emph{causal} temporal attention (\texttt{causal\_time\_attn=True}). It takes 5 context frames and predicts 1 future frame per rollout step. 

\textbf{Detailed short-horizon predictor.}
This predictor operates at 10 Hz and is conditioned on the abstract predictor's predicted latent. Temporal attention is \emph{non-causal}. Conditioning is injected spatially per-frame via small non-zero-initialized linear projections and a learned frame-position MLP, ensuring the pretrained STDiT backbone is minimally perturbed at initialization. 
Each block injects the abstract predictor's signal through dense adaptive layer normalization (AdaLN), predicting per-token scale and shift parameters from the spatially aligned abstract latent.

\textbf{Inference Protocol.}
At inference, generation proceeds in a cascaded fashion. The abstract predictor takes 5 context frames and generates 1 frame at 2\,Hz using 15 NFE. Its predicted abstract latent serves as the subgoal that conditions the detail predictor. The detail predictor then takes 5 context frames and generates the next 5 frames at 10\,Hz using 30 NFE, conditioned on the predicted abstract latent. During training, the detail predictor is conditioned on ground-truth abstract latents, whereas at inference it operates on the predicted latents from the abstract predictor. Rollouts are performed auto-regressively. The abstract predictor advances with a sliding window of 1 frame, operating over a 6-frame window (3\,seconds) at 2\,Hz. The detail predictor advances with a sliding window of 5 frames, operating over a 10-frame window (1\,second) at 10\,Hz.

\subsection{Probing details}~\label{app:linear_probe_details}

\paragraph{Semantic Segmentation}
To evaluate the quality of representations learned by the Stage-2 predictor model, we train a linear probe for semantic segmentation on the Cityscapes fine-annotation benchmark (19 classes) using the model's frozen intermediate features. 
Input images are normalized to [$-1$, $1$] and encoded through the frozen backbone. During probing, context is dropped (set to null) and the input image is tiled across all predicted-frame slots, with features extracted from the last frame slot. 
Intermediate patch-token features are extracted from the 4th transformer block from the end for the main model and from the 2nd block from the end for ablation models, then reshaped from the flat patch sequence into a 2D spatial feature map. 
The linear probe head is a single 1×1 LazyConv2d that maps this feature map to 19-class logits, followed by bilinear upsampling to match the ground-truth mask resolution. 
Only the probe parameters are trained, the backbone is kept entirely frozen throughout. Training uses AdamW (lr=1e-3, wd=1e-4) with cross-entropy loss for 100 epochs. Performance is reported as mean Intersection-over-Union (mIoU) on the Cityscapes validation set.

\paragraph{Depth Estimation (Linear).}
To probe the geometry-awareness of the Stage-2 predictor model, we train a linear depth probe on the KITTI Eigen split following the DINOv2 linear depth evaluation protocol.
Input images are encoded through the frozen backbone; during probing, context is dropped (set to null) and the input image is passed as the sole target frame, with features extracted from the last frame slot. 
Intermediate patch-token features are extracted from the 4th transformer block from the end, then reshaped into a 2D spatial feature map. The linear probe head follows the DINOv2 Lin-1 design: a single 1×1 LazyConv2d producing logits over 256 uniformly-spaced depth bins covering [0, 80] m, preceded by a 4× bilinear upsample to recover spatial resolution, followed by a softmax and weighted sum over bin centers to yield a continuous depth prediction. At training time, ground-truth depth values are discretized into the same uniform bins and cross-entropy loss is minimized with AdamW (lr=1e-3, wd=1e-4) for 10 epochs; missing and out-of-range depth pixels are excluded from the loss. Only the probe parameters are trained; the backbone is kept entirely frozen. Performance is reported as RMSE on the KITTI Eigen validation split.

\paragraph{Depth DPT probing protocol.}We measure how much metric depth information a frozen backbone encodes by training an identical DPT decoder on top of each backbone, with the backbone weights kept fixed. KITTI is split by drive so that the training, validation and test drives are disjoint, and the decoder is trained on the training drives only. All methods are evaluated on the same 2{,}060 test images, namely those for which the five preceding frames exist, so single-frame and multi-frame results are directly comparable. Multi-frame inputs consist of the target frame plus five past frames sampled at 2\,Hz or 5\,Hz. The decoder adds 18.9--20.0M trainable parameters, depending on backbone width. The parameter counts in Table~\ref{tab:kitti_dpt_probe_full} cover only the frozen backbone; for Orbis~2 this is the tokenizer encoder (86M) plus the predictor (513M). DA3 and VGGT were pretrained with dense depth supervision (marked \emph{Depth sup.}), whereas DINOv2, V-JEPA~2.1 and Orbis~2 were not.
Following common practice, we evaluate on pixels with valid ground truth inside the Garg crop and within 0--80\,m. Predictions are compared in metric scale, without median scaling. With $\hat{d}_i$ the predicted and $d_i$ the ground-truth depth of pixel $i$, and $N$ the number of valid pixels, we report the absolute relative error $\mathrm{AbsRel} = \frac{1}{N}\sum_i |\hat{d}_i - d_i| / d_i$, the root-mean-square error $\mathrm{RMSE} = \sqrt{\frac{1}{N}\sum_i (\hat{d}_i - d_i)^2}$ in metres, and the threshold accuracy $\delta_1$, the fraction of pixels with $\max(\hat{d}_i/d_i,\, d_i/\hat{d}_i) < 1.25$.

\begin{table}[t]
\centering
\small
\caption{Extended results for Table~\ref{tab:kitti_dpt_probe}: monocular depth on the drive-disjoint KITTI split from frozen backbones with a retrained DPT decoder. Depth sup.: backbone pretrained with dense depth supervision. RMSE in metres; best per block in bold.}
\label{tab:kitti_dpt_probe_full}
\setlength{\tabcolsep}{3pt}
\begin{tabular}{lcrllccc}
\toprule
Backbone & Depth sup. & Params & DPT & Input & AbsRel$\downarrow$ & RMSE$\downarrow$ & $\delta_1\uparrow$ \\
\midrule
\multicolumn{8}{l}{\textit{Single frame}} \\
DINOv2-B         &            & 87M   & retrained & 1 frame & 0.067 & 3.035 & 0.950 \\
V-JEPA 2.1 ViT-g &            & 1013M & retrained & 1 frame & 0.066 & 3.043 & 0.951 \\
DA3-Large        & \checkmark & 304M  & retrained & 1 frame & 0.071 & 3.205 & 0.940 \\
DA3-Giant        & \checkmark & 1136M & retrained & 1 frame & 0.067 & 3.037 & 0.954 \\
VGGT-1B          & \checkmark & 909M  & retrained & 1 frame & \textbf{0.060} & \textbf{2.875} & \textbf{0.961} \\
\textbf{Orbis 2 (ours)}          &            & 599M  & retrained & 1 frame & 0.065 & 3.016 & 0.953 \\
\midrule
\multicolumn{8}{l}{\textit{Five past frames + target}} \\
V-JEPA 2.1 ViT-g &            & 1013M & retrained & 6 @ 2\,Hz & 0.069 & 3.067 & 0.949 \\
V-JEPA 2.1 ViT-g &            & 1013M & retrained & 6 @ 5\,Hz & 0.066 & 2.999 & 0.955 \\
DA3-Large        & \checkmark & 304M  & retrained & 6 @ 2\,Hz & 0.061 & 2.824 & 0.968 \\
DA3-Giant        & \checkmark & 1136M & retrained & 6 @ 2\,Hz & \textbf{0.049} & \textbf{2.516} & 0.973 \\
VGGT-1B          & \checkmark & 909M  & retrained & 6 @ 2\,Hz & 0.053 & 2.573 & \textbf{0.977} \\
\textbf{Orbis 2 (ours)}          &            & 599M  & retrained & 6 @ 2\,Hz & 0.060 & 2.791 & 0.966 \\
\bottomrule
\end{tabular}
\end{table}

\begin{figure*}[h!]
    \centering
    \begin{subfigure}[b]{0.48\textwidth}
        \centering
        \includegraphics[width=0.9\textwidth]{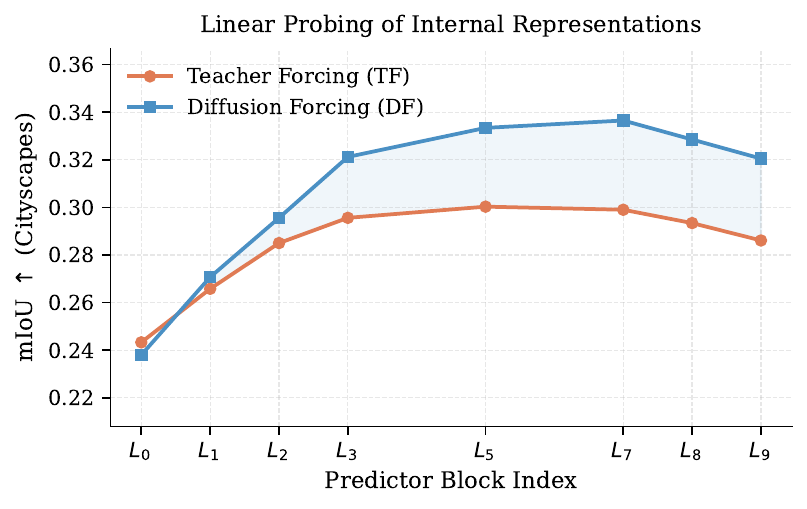}
        \caption{Semantic segmentation (mIoU $\uparrow$) on Cityscapes.}
        \label{fig:probing_miou}
    \end{subfigure}
    \hfill
    \begin{subfigure}[b]{0.48\textwidth}
        \centering
        \includegraphics[width=0.9\textwidth]{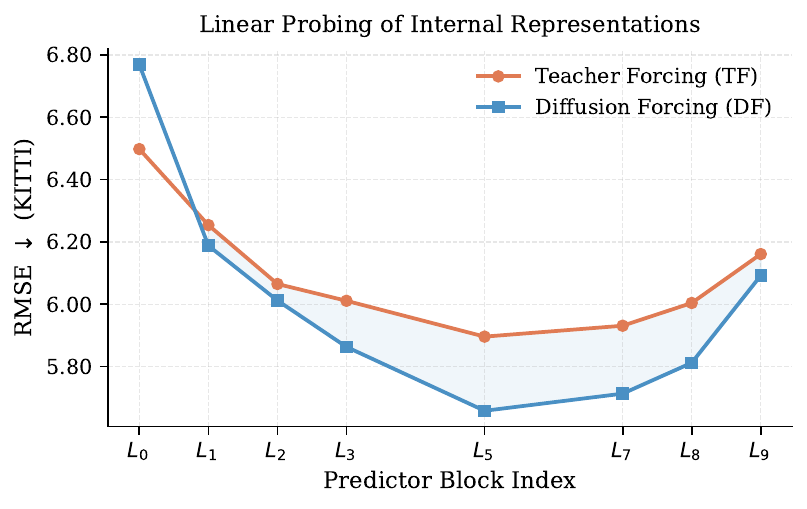}
        \caption{Depth estimation (RMSE $\downarrow$) on KITTI.}
        \label{fig:probing_rmse}
    \end{subfigure}
    \caption{Representation quality per predictor block via linear probing for the predictor trained on detail latents. Diffusion forcing consistently improves intermediate representations on both semantic segmentation and depth estimation.}
    \label{fig:linear_probing}
\end{figure*}

\paragraph{Per-block representation quality.}
Figure~\ref{fig:linear_probing} reports linear probing performance at each block of the predictor trained on detail latents, for the diffusion forcing and teacher forcing objectives discussed in \ref{sec:df_tf}. Diffusion forcing improves representation quality across the whole depth of the model on both probing tasks, not only in the final blocks.

\clearpage
\newpage

\end{document}